\documentclass[runningheads]{llncs}
\pdfoutput=1
\usepackage{graphicx}

\usepackage{tikz}
\usepackage{comment}
\usepackage{amsmath,amssymb} 
\usepackage{color}

\usepackage[accsupp]{axessibility}  

\begin{document}
\pagestyle{headings}
\mainmatter
\def\ECCVSubNumber{4}  

\title{Transformers and CNNs both Beat Humans on SBIR} 

\titlerunning{Abbreviated paper title}
%
\author{Omar Seddati \and
St\'{e}phane Dupont \and
Sa\"{\i}d Mahmoudi \and
Thierry Dutoit}
\authorrunning{O. Seddati et al.}
%
\institute{ISIA Lab (UMONS)\\
\email{\{omar.seddati, st\'{e}phane.dupont, sa\"{\i}d.mahmoudi, thierry.dutoit\}@umons.ac.be}}
\maketitle

\begin{abstract}
Sketch-based image retrieval (SBIR) is the task of retrieving natural images (photos) that match the semantics and the spatial configuration of hand-drawn sketch queries. The universality of sketches extends the scope of possible applications and increases the demand for efficient SBIR solutions. In this paper, we study classic triplet based SBIR solutions and show that a persistent invariance to horizontal flip (even after model finetuning) is harming performance. To overcome this limitation, we propose several approaches and evaluate in depth each of them to check their effectiveness. Our main contributions are twofold: We propose and evaluate several intuitive modifications to build SBIR solutions with better flip equivariance. We show that vision transformers are more suited for the SBIR task, and that they outperform CNNs with a large margin. We carried out numerous experiments and introduce the first models to outperform human performance on a large scale SBIR benchmark (Sketchy). Our best model achieves a recall of 62.25\% (at $k = 1$) on the sketchy benchmark compared to previous state-of-the-art methods 46.2\%. 
\end{abstract}

\section{Introduction}
The possible applications for SBIR solutions are multiple and especially with the large availability of touchscreens everywhere. In addition, the universality of sketches and the efficiency of modern SBIR solutions that are able to deal with badly drawn sketches make the use of such solutions less burdensome. For example, a SBIR solution may be integrated in an e-commerce system, where the user draws a sketch to find a specific product. Sketching will offer a powerful solution to communicate details that goes beyond the category of the object. These details could go from the global product design to detailed patterns present on the product and with respect to their actual spatial configuration. It would be difficult to reach such level of details using a query-by-text paradigm. 

In order to build SBIR solutions, researchers have conducted extensive studies and proposed several solutions to manage the abstract nature of sketches and extract cross-domain features. In this work, we use the classic triplet training pipeline to train our models on the Sketchy database. Then, we conduct a qualitative inspection of the model’s errors and show that the model struggles with horizontal flipping. In multiple cases, when the photo matching the sketch query comes at the second position, the retrieved photos holding first, and second place share the same semantics and even spatial configuration if mirrored. Which indicates that the models keep preserving in somehow flip invariance even after finetuning. Despite the value of such property for classification problems, in the particular case of SBIR where we seek for a fine-grained matching, such property represents a major weakness because object pose is not represented well enough. To overcome this limitation, we propose several methods (custom mini-batch sampling, adapting pooling layer, using a vision transformer) and conduct for each an additional finetuning to assess if it is beneficial. During the whole study process, we use intermediate assumptions and experiments to draw conclusions and build step-by-step our final end-to-end enhanced SBIR solution. Through this study, we show that a simple modification to a triplet CNN based SBIR solution at the last pooling layer is enough to bring a significant improvement. In addition, we also show that a vision transformer does not share the limitation observed with CNNs, which makes it more suited for SBIR. 

The rest of the paper is organized as follows: we start with a brief review of different approaches used in the field of SBIR. Next, we present the method that we selected as a baseline, and we proceed to the study and the evaluation of the different approaches that propose to enhance existing solutions. 

\section{Related Work }
In computer vision, it has been a few years since supervised learning with CNN achieves state-of-the-art results \cite{2,3,4}. CNNs are known for their ability to learn to extract relevant features directly from pixels without requiring classic feature extraction approaches. This feature makes them a convenient and powerful tool for multiple computer vision tasks. In addition, compared to hand crafted features (shallow features), CNN features reach higher performance \cite{5,6,7} when it comes to generating representations for tasks like content-based image retrieval (CBIR). Sketch recognition and SBIR have also benefited from the power of CNN and outperformed previous solutions based on hand crafted features (e.g. \cite{23,24,25,26,27}). In several CNN based works \cite{11,10,9,8}, the authors use a CNN trained for sketch recognition to extract features and build a sketch-edge map matching solution (under the assumption that photos edge maps are visually closer to sketches). In \cite{11}, Qi et al. used a Siamese CNN architecture, for category-level SBIR (which means that the result is considered as correct if the retrieved photo belongs to the same category as the sketch query). In such architecture, two branches are used, one for the sketches and the other for the edge maps. During training, sketch-edge map pairs are passed to the model and a binary label indicates if the sketch and the edge map share the same category. Then the loss is computed and the model parameters are updated to extract better representations in the future. In a more general fashion, pair losses are used with the objective to learn to project inputs in a feature space where the distance between positive pairs (similar inputs) is smaller than a margin $m_{p}$, while the distance for negative pairs is larger than a second margin $m_{n}$. Using the same margin for all pairs constitutes a major disadvantage since it does not consider the variance of (dis)similarity between different pairs. This limitation is avoided when a Triplet CNN is used. A similar approach where inputs are presented as triplets. In each triplet we have a reference sample, a positive sample (similar to the reference) and a negative sample (dissimilar to the reference). During training, the model learns to project inputs into a space where a positive example is closer to the reference than a negative one. This approach based on relative distance measure gives the model the ability to manage arbitrary feature space distortions. This property makes Triplet CNN more suited for CBIR/SBIR applications and attracted a lot of attention \cite{12,13,14,15,16,17,18,19} in the last recent years. In \cite{12}, Bui et al. made several experiments with triplet CNN to explore weight sharing strategies and analyze its ability to generalize. In \cite{14}, the authors added a deformable CNN layer (changeable receptive field) to deal with sketch changes. In \cite{13}, in addition to the triplet loss, the authors experimented with a combination three loss functions (SoftMax loss, Spherical loss, and Center loss). In \cite{16,17,18,19}, the authors added attention modules to enhance fine information capture. In \cite{21}, Bhunia et al. addressed the problem related to the lack of annotated sketches. They proposed an approach where a photo-to-sketch generator (a sort of GAN) is used to automatically generate sketches for unlabeled photos. The generated synthetic sketch-photo pairs are then used as additional data to train a triplet CNN. In \cite{20}, the authors introduced the quadruplet networks, a new architecture for SBIR. It uses an additional branch compared to a triplet architecture to ensure that for each reference, there are two negatives. The first one belongs to the same object category, while the other is from a different category. The goal of such architectures is to encode semantic information in a similar fashion as what triplet does for local information. In \cite{15}, Wang et al. proposed a sophisticated SBIR solution where textual descriptions (collected from MSCOCO and Flickr30k databases) are used as additional input to the pipeline to reduce the gap between sketches and images. They propose a three-stage solution, where a deep multimodal embedding model is used to extract features for the three modalities. Then, a candidate selection based on the extracted embeddings is applied. In the final stage, a triplet is used for ranking optimization. In \cite{22}, a cross-modal variational autoencoder is proposed to disentangle sketch into semantic content (an information shared with the corresponding photo) and sketcher style to build a style-agnostic model. 

All these works have brought various adaptations and new solutions for SBIR. In this work, we take a closer look to the errors that occurs when using a triplet CNN approach and how we can overcome its limitations.
\section{Methodology and Experiments }
In order to build an efficient SBIR system, we need a solution able to analyze in depth both sketches and images to generate representations that seize the semantics and spatial configurations. Before we proceed to the core of this study, let introduce how a triplet network works and what makes it suitable for the SBIR task. 
\subsection{Triplet Network}
In this kind of architecture, we generally have three copies of the same network with shared parameters. The input is in the form of Anchor, Positive and Negative. In the context of SBIR, if we use photos as anchors, the goal will be minimizing the distance between an anchor (a photo) and a positive (a sketch corresponding to the anchor), while maximizing the distance between this anchor and a negative (a sketch not corresponding to the anchor). In order to do that, we train our triplet network with the following loss function:
\begin{equation} \label{eq:2}
\sum_{i}^{N}[ \left \| f(x_{i}^{a}) - f(x_{i}^{p}) \right \|_{2}^{2} - \left \| f(x_{i}^{a}) - f(x_{i}^{n}) \right \|_{2}^{2} + \alpha ]_{+}
\end{equation}
where :\\
$x_{i}^{a}$: the feature vector of the anchor photo.\\
$x_{i}^{p}$: the feature vector of the positive sketch.\\
$x_{i}^{n}$: the feature vector of the negative sketch.\\
$\alpha$: by using this margin we try to ensure a minimal margin between the negative pair and positive pair distances.
\subsection{The Sketchy Dataset}
%
%
\begin{figure*}
\centering
  \includegraphics[height=3cm]{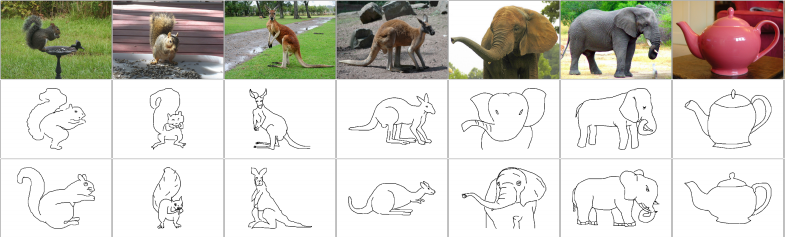}
\caption{Samples from the sketchy database \cite{23}.}
\label{fig:1}       
\end{figure*}
For the whole study, we will use the Sketchy benchmark \cite{1}. It is a large-scale collection for SBIR. It contains 125 categories and 75,471 sketches of 12,500 objects (some samples are shown in Figure \ref{fig:1}). The crowd workers are asked to sketch particular photographic objects in order to give a fine-grained association between sketches and photos. For each category we have 100 images and at least 5 sketches per image. We followed the guidelines of the authors \cite{1} and used the provided testset list to split data into training and testset. 90\% of the data are used for the training and the rest are used for test.
\subsection{Baseline Models and Practical Details }
We began our study by training triplet CNNs with three different architectures \cite{34} (ResNet18, ResNet34, ResNet50) on the Sketchy dataset. We use the pre-trained (pre-trained on ImageNet \cite{40}) versions provided by the torchvision library. We select the last pooling layer (adaptive average pooling layer) to extract features (in this work, we do not normalize the extracted embeddings and we use the Euclidean distance), and we keep the classification branch (we change the number of neurons of the fc layer to correspond to the 125 categories of Sketchy) to build a multi-task triplet network. During training, we use two different models, one for sketches and the other for photos. We train each architecture for 100 epochs using the Adam optimizer \cite{31} and we set the initial learning rate to $lr = 10^{-4}$. The learning rate is changed to $lr = 10^{-5}$ after 30 epochs. The batch-size (bs) and the triplet margin (m) remain the same for the whole study ($bs = 128$ and $m = 3$). The best baseline models' performances are reported on the first line of Table \ref{table:report1}. We can notice that replacing a ResNet18 with a ResNet34 improves significantly the results. But the opposite happens when replacing a ResNet34 with a ResNet50. This may come from the fact that in ResNet50 the basic blocks are replaced with bottleneck blocks to reduce memory usage. In these blocks, pointwise convolution is used to reduce the number of channels before $3\times 3$ convolutions. And the loss of performance might be a side effect of this form of compression. 
\setlength{\tabcolsep}{4pt}
\begin{table}
\begin{center}
\caption{The top results ($recall @1$) achieved on the Sketchy benchmark using different architectures and methods.}
\label{table:report1}
\begin{tabular}{llll}
\hline\noalign{\smallskip}
Method & ResNet18 & ResNet34 & ResNet50 \\
\noalign{\smallskip}
\hline
\noalign{\smallskip}
Baseline & 52.98 & 56.10 & 54.89 \\
Finetuning: Continue & 52.98 & 56.10 & 55.08 \\
Finetuning: Flip sampling & 53.78 & 57.19 & 55.91 \\
Finetuning: Category sampling & 53.83 & 56.92 & 56.24 \\
Finetuning: Flip + Category sampling & 53.61 & 57.43 & 56.29 \\
Finetuning: Pooling $2\times 2$ & 55.10 & 58.23 & 58.37 \\
\hline
\end{tabular}
\end{center}
\end{table}
\setlength{\tabcolsep}{1.4pt}
\subsection{Qualitative Inspection}
%
%
\begin{figure*}
\centering
  \includegraphics[height=6cm]{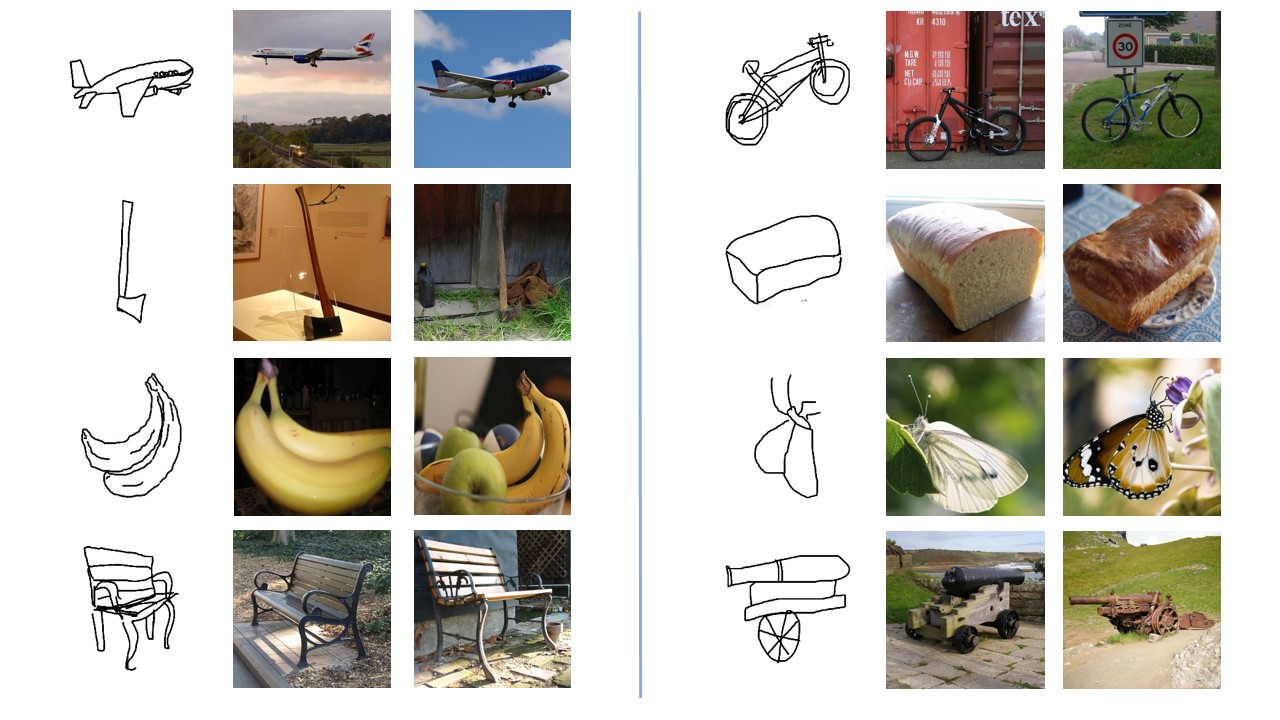}
\caption{Some examples showing that ResNet18 baseline still struggle in some cases with horizontal flip.}
\label{fig:r18bl_flip}       
\end{figure*}
When using the Sketchy benchmark, the recall metric, and especially the $recall @1$ is used to evaluate and compare different solutions. Since our models have reached an exceptionally high level of performance (human performance $recall @1$ is 54.27\% as reported in \cite{1}) and in order to see if there is much room for improvement, we did also compute the $ recall @2$ (the results are reported in Table \ref{table:report2}). The fact that the chances to retrieve the wanted photo increase significantly when we retrieve more than one image, motivated us to look closer to the errors made by the models, hoping to find a way to push $recall @1$ to get closer to $recall @2$. During this qualitative inspection, we noticed that in many cases, both, the first and second retrieved images exhibit an obvious similarity, but the models seem to struggle with the orientation as shown in figure \ref{fig:r18bl_flip}. Which indicates that the models keep preserving in somehow flip invariance even after finetuning. Despite the value of such property for classification problems, in the particular case of SBIR where we seek for a fine-grained matching, such property represent a major weakness because object pose is not represented well enough. In order to overcome this limitation, in the next subsections, we explore several approaches and show which of them bring a significant improvement. 

In addition to the flip invariance problem, we have also noticed during the aforementioned qualitative inspection that in some cases (in $18.5\%$ of the cases for the ResNet18 baseline model) the category of the image retrieved at the first position does not match the query’s category. This is another aspect of the model that we thought that we should be able to improve through a better mini-batch sampling during the training phase. More details about the approach that we propose will be presented in the subsection \ref{sec:catge_sampling}. 
\setlength{\tabcolsep}{4pt}
\begin{table}
\begin{center}
\caption{The top results ($recall @2$) achieved on the Sketchy benchmark using different architectures and methods.}
\label{table:report2}
\begin{tabular}{llll}
\hline\noalign{\smallskip}
Method & ResNet18 & ResNet34 & ResNet50 \\
\noalign{\smallskip}
\hline
\noalign{\smallskip}
Baseline & 69.24 & 73.75 & 71.88 \\
\hline
\end{tabular}
\end{center}
\end{table}
\setlength{\tabcolsep}{1.4pt}
\subsection{Flip Sampling}
Before moving forward to the evaluation of the different approaches we have considered in this work, we have begun with finetuning the baseline models for an additional 100 epochs on the same task. For each architecture (e.g., ResNet18), we reload the model achieving the best performance, we set the learning rate to $lr = 10^{-6}$ and we continue the training following the same procedure. This provides us a fair ground for future comparisons and improvements analysis. The results are shown in the second line \footnote{If the value at the second line is the same as the corresponding one from the first line, this means that the additional 100 epochs finetuning did not brought any improvement.} of Table \ref{table:report1}. 

Flipping the inputs, simultaneously the sketch and the photo, is the only data augmentation technique used during the classic training procedure adopted in this work. For the flip invariance problem, the first intuitive approach that we came out with was the adaptation of the mini-batch sampler. We modify the latter to include both, the flipped and non-flipped versions of some samples in the same mini-batch. With this modification, we expect the model to focus beyond the global semantics and pay more attention to the spatial configuration systematically within each mini-batch. As reported in the third line of Table \ref{table:report1}, we can notice that the proposed approach brings a substantial improvement. 
\subsection{Category Sampling }
\label{sec:catge_sampling}
Like the solution described in the previous subsection, here we modify the mini-batch sampler to make sure that inside the same mini-batch, multiple samples with the same category appears a random number of times (a random number is selected between 2 and 5). The aim of this modification is to build harder training samples (it is defensible to think that it is harder to differentiate two objects from the same category compared to objects from different categories) and strengthen the model’s ability to manage intra-categorical information. In order to evaluate the improvement brought with our approach, we compare the evolution of the number of times that the first retrieved photo is from a category not matching the one of the query (while the second photo is the wanted image). In Table \ref{table:sampler_improve} we report the percentage with which this number drops for each architecture when using the proposed approach. For example, if the number of errors for the baseline model is 100, and this number drops to 90 after using category sampling, the reported percentage will be 10\%.
\setlength{\tabcolsep}{4pt}
\begin{table}
\begin{center}
\caption{The percentage of improvement noticed when using our category sampling approach.}
\label{table:sampler_improve}
\begin{tabular}{llll}
\hline\noalign{\smallskip}
Method & ResNet18 & ResNet34 & ResNet50 \\
\noalign{\smallskip}
\hline
\noalign{\smallskip}
Baseline & 8.95 & 6.47 & 5.49 \\
\hline
\end{tabular}
\end{center}
\end{table}
\setlength{\tabcolsep}{1.4pt}
\subsection{Task Specific Pooling}
\label{sec_pool}
Here, we present an approach that differs fundamentally from the previous ones. We propose to introduce a tiny modification to the model's architecture in order to extract richer embeddings preserving more information about spatial configuration. We keep using the output of the original average pooling layer for the classification branch of our models. And we add an additional $2\times 2$ adaptive average pooling for the embedding branch to replace the previous one. The motivation behind such modification, is that with the original $1\times 1$ average pooling, the only option left to the model to encode spatial information is through channels. So, we continue to use it for the classification branch, since we think that it is enough for this task. But for the other branch, we pass the outputs feature maps of the last convolutional layer through a new $2\times 2$ adaptive average pooling (which produces 4 times longer vectors than before) to form our new embeddings. The rest of the training follows the same procedure. In the last line of Table \ref{table:report1}, we show how this tiny modification brings a significant improvement to the different architectures compared to the baseline models. 
\subsection{Vision Transformer }
\label{sec_vt}
Since a few years now, CNNs have been singled out for their lack of relative position encoding for different features because of the use of pooling layers \cite{sabour2017dynamic}. Vision Transformers (VT) on the other hand, are known for their ability to track long-range dependencies within an image. Thanks to self-attention layers, VT models are able to efficiently aggregate global information. This property drew our attention to VT models and their usage for the SBIR task. We have looked for other works on the field of SBIR using VT models, and we were surprised to find out that only three works \cite{gupta2022zero,ribeiro2020sketchformer,tian2022tvt} have used VT. In addition, in two of them \cite{gupta2022zero,tian2022tvt}, the proposed solutions are for Zero-shot SBIR. While in the other \cite{ribeiro2020sketchformer}, no one of the common SBIR benchmarks is used to evaluate the performance. Therefore, we have followed the same approach as previously, we start with training a baseline VT model. To this end, we use the Hugging face library \cite{wolf-etal-2020-transformers} and the pre-trained VT model under the name “openai/clip-vit-base-patch32" based on the work of Radford et al. \cite{radford2021learning}. The initial learning rate we used to train the baseline VT model is $lr = 10^{-5}$.  We have also finetuned the baseline VT model for an additional 100 epochs (with $lr = 10^{-6}$) for each of the proposed approaches (except for the pooling-based one). The outcomes of the different experiments with the VT model are listed in Table \ref{table:report1withvt} (side by side with those of ResNet for ease of comparison). We can see that the VT model outperforms by a large margin ResNets even when the pooling approach is used. We can also notice that in the case of VT models, the proposed approaches do not bring a significant improvement \footnote{In Table \ref{table:report1withvt}, we show that finetuning the VT model just with setting the learning rate to $lr = 10^{-6}$ is enough to reach a $recall @1$ of 61.74\%. Adding the approaches that we proposed and validated previously for ResNets, does not seem to provide additional benefit.}. Which may suggest that as expected, VT models do not share the same limitations as ResNets and makes them more suited for the SBIR task. In Table \ref{table:bench_res}, we compare our best models results with others from different papers and show that this study enabled us to outperform existing approaches with a huge margin, and even exceed for the first-time human performance on a large-scale SBIR benchmark.
\setlength{\tabcolsep}{4pt}
\begin{table}
\begin{center}
\caption{The top results ($recall @1$) achieved on the Sketchy benchmark using different architectures, including the VT model.}
\label{table:report1withvt}
\begin{tabular}{lllll}
\hline\noalign{\smallskip}
Method & ResNet18 & ResNet34 & ResNet50 & \bf VT\\
\noalign{\smallskip}
\hline
\noalign{\smallskip}
Baseline & 52.98 & 56.10 & 54.89 & \bf 56.24 \\
Finetuning: Continue & 52.98 & 56.10 & 55.08  & \bf 61.78 \\
Finetuning: Flip sampling & 53.78 & 57.19 & 55.91 & \bf 60.56 \\
Finetuning: Category sampling & 53.83 & 56.92 & 56.24 & \bf 62.08 \\
Finetuning: Flip + Category sampling & 53.61 & 57.43 & 56.29 & \bf 62.25 \\
Finetuning: Pooling $2\times 2$ & 55.10 & 58.23 & 58.37 & - \\
\hline
\end{tabular}
\end{center}
\end{table}
\setlength{\tabcolsep}{1.4pt}
%
%
\setlength{\tabcolsep}{4pt}
\begin{table}
\begin{center}
\caption{Comparison with some SBIR solutions on Sketchy \cite{1}.}
\label{table:bench_res}
\begin{tabular}{ll}
\hline\noalign{\smallskip}
Methods&Recall@1 \\
\noalign{\smallskip}
\hline
\noalign{\smallskip}
    Chance \cite{1}&0.01\%\\
    Sketch me that shoe \cite{41}&25.87\%\\
    Siamese Network \cite{1}&27.36\%\\
    Triplet Network \cite{1}&37.10\%\\
    Quadruplet\_MT \cite{20}&38.21\%\\
    DCCRM(S+I) \cite{15}&40.16\%\\
    DeepTCNet \cite{42}&40.81\%\\
    Triplet attention \cite{19}&41.66\%\\
    Quadruplet\_MT\_v2 \cite{20}&42.16\%\\
    DCCRM(S+I+D) \cite{15}&46.20\%\\
    Human \cite{1}&54.27\%\\
    \bf Ours (ResNet18 \ref{sec_pool})&\bf 55.10\%\\
    \bf Ours (ResNet34 \ref{sec_pool})&\bf 58.23\%\\
    \bf Ours (ResNet50 \ref{sec_pool})&\bf 58.37\%\\
    \bf Ours (VT \ref{sec_vt})&\bf 62.25\%\\
\hline
\end{tabular}
\end{center}
\end{table}
\setlength{\tabcolsep}{1.4pt}
\section{Conclusions}
In this work, we have selected a classic pipeline for building SBIR solutions. We started with training multiple architectures in a classic fashion to serve as a baseline for future comparisons. Then, we have conducted a qualitative inspection of the errors made by the model on the testset. Based on the observations, we identified a few limitations that we can address in an effort to improve existing solutions. To this aim, we proposed and motivated several approaches that we explored and evaluated on a large-scale SBIR benchmark. We have been able to show that the different approaches proposed implying the use of adapted mini batch sampling bring a substantial improvement. Furthermore, we demonstrate that with a tiny modification to the architecture at the pooling layer results on significant improvements. In addition, we did also show that VT models represent better candidates than CNNs for tasks with SBIR needs and how they surpassed remarkably the latter, and even human performance on the Sketchy benchmark.



\clearpage
%
%
\bibliographystyle{splncs04}
\bibliography{eccv}
\end{document}